\documentclass{llncs}
\usepackage[utf8]{inputenc}
\usepackage{amssymb,amsmath,array}
\usepackage{url}
\usepackage{graphicx}

\usepackage[section]{algorithm}
\floatname{algorithm}{Algorithme}
\usepackage{algorithmicx}
\graphicspath{{Images/}}
\begin{document}

\title{Lasso based feature selection for malaria risk exposure prediction.}

\author{Bienvenue Kouway\`e \inst{1,2,3}  \and No\"el Fonton  \inst{2,3} \and Fabrice Rossi \inst{1}}

\institute{Universit\'e Paris1 Panth\'eon-Sorbonne  - Laboratoire SAMM 
70, rue de Tolbiac 75013  - France
\and  Universit\'e d'Abomey-Calavi, International Chair in Mathmatical 
Physic and Applications (ICMP:UNESCO-Chair), Abomey-Calavi, B\'enin
\and Universit\'e d'Abomey-Calavi, Laboratoire d'étude et de
recherche en statistique appliquée et biométrie (LERSAB)}

\maketitle

\begin{abstract}
In life sciences,  the experts generally use  empirical knowledge 
to recode variables, choose interactions and  perform   selection 
by classical  approach.  The aim of this work is to perform automatic learning
algorithm for variables selection which can lead to  know 
if  experts can be help in they decision or simply  replaced by the machine  
and improve they  knowledge and 
results. The Lasso method can  detect the optimal subset of variables for estimation and 
prediction under some conditions. In this paper,  we propose  a novel  approach 
which  uses  automatically all  variables available and all interactions.
By a double   cross-validation combine with 
Lasso, we select a best subset of variables and with GLM through  
a simple cross-validation perform  predictions. The algorithm  assures the stability and the 
the consistency of estimators.
\end{abstract}

\section{Introduction}
Malaria  is endemic in developing countries, mainly in sub-Saharan Africa. 
It  is the main cause of mortality  especially for children under five years
of age  in Africa \cite{WHO}.
Generally, cohort studies take place in endemic areas for  characterizing  the malaria risk. 
These cohorts  studies are on newborn babies and pregnant women.
They are  introduced    to know about the immunity of newborn face to malaria  and the 
construction of this immunity. They also 
help to know  the determinants implicated in the  appearance 
of first malaria infections on the newborn.
The distribution of the  main 
vector of malaria (anopheles) and the exposure to malaria risk are  spatiotemporal and different 
at small scale (house level) \cite{cottrell2012malaria}. 
Generally the experts in medicine and epidemiology, use they empirical knowledge 
on phenomenon in the process of data analysis. Some variables are automatically recoded
because in numeric form, they are not interpretable \cite{cottrell2012malaria} 
and interactions are chosen. 
Classical variables selection methods are wrapper (forward, backward, stepwise, etc.), embedded,  filter and ranking. 
All details on theses methods are described in \cite{Guyon03anintroduction}. The goal of  wrapper method
is to select  subset of variables with a low prediction error. 
The wrapper algorithm is improved by Structural 
wrapper to obtain a sequence of nested subset of feature for optimality is developed in
 \cite{DBLP:conf/esann/Bontempi05}. 
In  practice,  classical method of features selection are practically unfeasible in high dimension
because  the number of features subsets given by  ($2^p$) where $p$ is the number of features,  increases. 
The Lasso method proposed by Tibshirani \cite{tibshirani1996regression} is a regularized estimation 
approach for regression models that 
 uses an $L_{1}$-norm and constrains the regression coefficients.  The Lasso method minimize the likelihood $L(\beta)$ or the
 log-likelihood $l(\beta)$.
 The results of this method  is that all coefficients are shrunken 
 toward zero and some are set exactly to zero. Many studies \cite{efron2004least}, \cite{goeman2010l1},
 \cite{Osborne1}, \cite{zou2005regularization} have improved
 the original Lasso method. 
 The Generalized Linear Model (GLM) combined with the Lasso, 
 penalize the log-likelihood of the GLM and constrains the $L_{1}$ or $L_{2}$ or ($L_{1}$+$L_{2}$)-norm 
 of the regression coefficients to be inferior to
 some parameter known as tuning or regularizing parameter \cite{goeman2010l1}, \cite{zou2005regularization}. 
 The usage of penalization technique to select variables in generalized
 models is at embryonic stage.
 Most of the algorithms implemented in our work are based on \cite{Friedman2015glmnet}, \cite{goeman2010l1}, \cite{zou2005regularization}. 
According to the nature of the target variable, family of models used for feature selection, estimation and prediction are generally
linear models, generalized models, mixed models, generalized mixed models, multilevel modeling \cite{cottrell2012malaria}.
It is also well known that cross-validation may lead to overfitting\cite{cross-validation1} and one alternative solution 
is $precentile-cv$ \cite{Ng97preventing"overfitting"}.
The results will be compared to the results of those of the reference method 
(B-GLM) which uses a backward procedure combine with a GLM \cite{cottrell2012malaria}
\section{Methodology}

\subsection{Data collection and variables}
The study area was conducted in Tori-Bossito a district of Republic of B\'enin, 
between July 2007 and July 2009. The study area (season, vegetation), the  methodology
of mosquito collection and identfication, the environnement and behavioral data are described in \cite{cottrell2012malaria}.
The dependent variable was the number of
Anopheles collected in a house over the three nights of each
catch, and the explanatory variables were the environmental
factors, i.e. the mean rainfall between two catches (Rainfall), the number of rainy days in the ten days 
before the catch (RainyDN10), the season during which the catch was carried out (Season), the type of 
soil 100 meters around the house (Soil),
the presence of constructions within 100 meters of the house (Works),
the presence of abandoned tools within 100 meters of the house (Tools), the
presence of a watercourse within 500 meters of the house (Watercourse), the type of vegetation 100 meters around the house (NDVI), the type of roof (Roof), the number of
windows (Windows), the ownership of bed nets (Bed nets), the use of insect 
repellent and the number of inhabitants in the house (Repellent), see more details in  \cite{cottrell2012malaria},
the number of rainy days during the three days of one survey (RainyDN).
In the previous  work \cite{cottrell2012malaria}, a  second type of  variables are obtained by recoding the original explanatory variables one based
of the knowledge of experts in entomology and medicine. The  Original and recoded variables are described in tables
(\ref{tab:Originales variables},\ref{tab:Recoded variables}).\\
Four type    of combination of these variables have been used. The firs is only the original variables, 
the second the original variables with village as fixed effect, 
the third is recoded variables and the last is 
the recoded variables with village as fixed effect.
\subsection{Models}
The algorithm implemented  uses  GLM-Lasso to detect the entire paths 
of the variables, in  order to detect all changes in the process of regularization.
This regularization  consist on  penalizing the likelihood of the GLMs by adding a penalty term 
\begin{equation}
 \mathcal{P}(\lambda) =\lambda  \sum_{i=1}^{p} {\lvert\beta_{i}\lvert}
\end{equation}
with $\lambda \geq 0$.\\
Then the log-likelihood penalized is :
\begin{equation}
 \label{Eq4}
l_{pen}(\beta| Y) = l_{GLM}(\beta| Y) + \mathcal{P}( \lambda)
\end{equation}
The penalty problem is reduce to :
\begin{equation}\label{lasso}
 \hat{\beta} = Arg\max_{\beta}{\left[l_{GLM}(\beta| Y)+ \mathcal{P}( \lambda)\right]}
\end{equation}
The choice of the regularizing parameter lambda is done by minimizing a score function. In  practice, this equation doesn't have exact  numerical  solution. It can be used the combination of Laplace approximation, the 
Newton-Raphson method or the Fisher scoring to solve it. This procedure is used at each learning step.\\
One of the power of the Lasso is to shrink some coefficients  toward to zero and the other to exactly zero. When the 
regularizing parameter $\lambda$ exceeds certain threshold $(\lambda_{max})$, the intercept is the only non-zero coefficient \cite{tibshirani1996regression}. 
For two different  values of the regularizing parameter $\lambda$ and $\lambda'$ inferior to $\lambda_{max}$,
let  $\beta$ and $\beta'$ be  the vectors of
fixed coefficients respectively associated to $\lambda$ and $\lambda'$, $\beta \neq \beta'$
\cite{efron2004least,tibshirani1996regression}. Then let us  define the function
 \begin{displaymath}
      Q:  
      \begin{array}{l}
              \lambda_{i} \longmapsto \hat{\beta}_{i}\cr
              \end{array}
      \end{displaymath}
$\lambda_{i} \;\in \; \left[0, \lambda_{max}\right] $,  $\hat{\beta}_{i}$ is the vector of coefficients 
$\hat{\beta}_{i} = (\hat{\beta}_{i1}, \hat{\beta}_{i2},..,\hat{\beta}_{ip})$,
$p$ is the number of coefficients in the model. 
The parameter  $\lambda$ is considered as discreet then  $ \lambda_{i} \;\in\; \{\lambda_{0}, \lambda_{1},...,\lambda_{max}  \}$.
Because the lasso coefficients are biased,
GLM is used to debiased estimators and makes predictions.
Under matrix shape, GLM model is 
 \begin{equation}
g[E(Y|\beta)]=X\beta 
\end{equation}
where $(Y |, \beta)$ follow a Poisson distribution of parameter  $E(Y | \beta)$, \\
$n$\, is the number of observations, $ X$  the $n \times \,(p+1)$-dimension matrix of 
co-variables (environmental variables), 
$\beta$ is a $(p+1)$-vector of fixed parameters  including the intercept, 
$Y$ is the vector  of   the target variable.
\begin{equation}
 (Y|X=x) \sim \mathcal{P}(\lambda); \;\;  \mbox{where}  \;\; \lambda = e^{x\beta}
\end{equation}
Then 
\begin{equation}
 \mathbb{P} ((Y=y_i|X=x)) = \frac{e^{(x\beta)^{y_i}}}{(y_i)!} \times e^{-e^{x\beta}}
\end{equation}
With 
$Z_i=(Y=y_i|X=x)$,
the likelihood on $n$ observations can defined as
 \begin{equation}
  L(Z_1, \ldots, Z_n) = \prod_ {i=1}^{n} \frac{e^{(x\beta)^{y_i}}}{(y_i)!} \times e^{-e^{x\beta}}
 \end{equation}
And the log-likelihood is
\begin{eqnarray}
  \mathcal{L}(Z_1, \ldots, Z_n) &= &\log\left( \prod_ {i=1}^{n} \frac{e^{(x\beta)^{y_i}}}{(y_i)!} \times e^{-e^{x\beta}}\right)\cr
  &=& -\sum^{n}_ {i=1}\log((y_i)!) + \sum^{n}_ {i=1} y_i(x\beta) - e^{(x\beta)}
 \end{eqnarray}
 $y_i$ don't depend on $\lambda$ then 
 \begin{equation}
  \mathcal{L}(Z_1, \ldots, Z_n) =Cste + \sum^{n}_ {i=1} y_i(x\beta) - e^{(x\beta)}\quad \mbox{where}\quad  Cste =-\sum^{n}_ {i=1}\log((y_i)!)
 \end{equation}

\subsection{Leave One Level Out Double Cross-Validation (LOLO-DCV)}
This algorithm is a double cross-validation with  two  levels.
Its aim is to compute a  second cross-validation (CV2)  for  prediction at each 
step of learning  of a first cross-validation (CV1). The predictors obtained with 
(CV2) are consistent for prediction on the test set for (CV1).
This algorithm run like described in Algorithm (\ref{LOLO_DCV_Algorithme}).
 \begin{algorithm}{}
 \caption{LOLO-DCV}
 \begin{enumerate}
  \item 
The data are separated  in  $N$-folds
 \item A each step of the first level
 \begin{enumerate}
\item The folds are regrouped in two part : $E_A$ and $E_T$, 
$E_A$ : the learning set which 
contained the observations of $(N-1)$-folds,\\ $E_T$ : the test set, contained the observations of the last fold. 
\item Hold-out $E_T$
\item \label{CV1} The second level of cross-validation
\begin{enumerate}
\item    A full  cross validation is compute  on $E_A$ 
  \item  The two  regularizing parameters  "lambda.min" and lambda.1se" are got back.
  \item The coefficients  of  actives variables (variables with non-zero coefficient) 
  associated to these two parameters are debiased
    \item Prediction are performed using a glm model on $E_T$
  \item The presence $\mathcal{P}(X_i)$ of each variable is determined
  \end{enumerate}
\end{enumerate}
\item The step (\ref{CV1}) is repeated  until predictions is performed for all observations.
\end{enumerate}
\label{LOLO_DCV_Algorithme}
\end{algorithm}
LOLO-DCV uses a score of cross validation defined like:
\begin{equation}
  Score( \lambda_{i}) =Deviance(\lambda_i)  = 2\times \left( \mathcal{L}_{(sat)} - \mathcal{L}_{(\lambda_i)}\right)
  \label{Score_Equation}
\end{equation}
Where $\mathcal{L}_{(sat)}$is the log-likelihood of saturated model  and $\mathcal{L}_{(\lambda_i)}$ 
the log-likelihood of  the concerned  model. 
For   $\lambda=\lambda_{max}$ the model obtained is the null model which contain only the intercept and $\mathcal{L}_{(sat)}=0$,
the model which adjust perfectly data. Then
\begin{equation}
 Score( \lambda_{max})=Deviance(NULL)  =- 2\times \left(\mathcal{L}_{\lambda_{max}}\right)
\end{equation}
 \begin{eqnarray}
   Score( \lambda_{i})  & =& -2\times \left( \mathcal{L}_{(\lambda_i)}\right)\cr
   &=&-2\left[ \left(Cste + \sum^{n}_ {i=1} y_i(x\beta) - e^{(x\beta)}\right)\right], \;\;\; k \in \mathbb{R}\cr
   Score( \lambda_{i})&=& 2\left[ \left(  e^{(x\beta)} -\sum^{n}_ {i=1} y_i(x\beta) \right)\right] + K, \;\;\; K \in \mathbb{R}
 \end{eqnarray}
To minimize the score is reduced to minimize the quantity $\left(\sum^{n}_ {i=1} y_i(x\beta) - e^{(x\beta)}\right)$.
In gaussian case, the score used is the mean cross-validation error, a vector of length $\lambda$ \cite{Friedman2010regularization}.
Assume that :
\begin{equation}
 R= 1-\frac{Score( \lambda_{i})}{Score( \lambda_{max})}=1-\alpha
\end{equation}

then :\\
\begin{equation}
 Deviance(\lambda_i) = (1-R)\times Score( \lambda_{max}) 
 \label{deviance}
\end{equation}
We know that $\mathcal{L}_{(sat)} = 0$ then  $\alpha$ is a log-likelihood ratio between the concerned model and the null model.
The optimal value   $\lambda.min$ of  $\lambda$ is the one minimize the  $Score(.)$ function. 
\begin{equation}
 \lambda.min= Arg\min_{\lambda_{i}} [Score(\lambda_{i})]   
\end{equation}For all positive  value of $\lambda_{i}$,  the score exist and is finite.
it shows that the score of cross validation converge.
The optimal value of $\lambda_{0j}$ is given by  the  minimization of  the function $Score(.)$. 
\begin{equation}
 \lambda_{0j}= Arg\min_{\lambda_{i}} [Score(\lambda_{i})]   
\end{equation}
If $\lambda_{0j}=\lambda_q$ then the parameter $lambda.min$ is  $lambda.min=\lambda_q$ \cite{Friedman2015glmnet}, \cite{hastie2009element}.
Let  $Std(Score(.))$ the estimate of standard error of $Score(.)$.
Suppose
\begin{equation}
 \lambda_{0k} = Arg\min_{\lambda_{i}} [Score(\lambda_{i}) + Std(Score(\lambda_{i})]  
\end{equation}
If $\lambda_{0k}=\lambda_{m}$ then the parameter $lambda.1se$ is defined as\\
$lambda.1se=\lambda_m$ \cite{Friedman2015glmnet}, \cite{hastie2009element}.
\subsection{Prediction power and quality criteria}
During the computation of LOLO-DCV :  
\begin{enumerate}
 \item GLM-Lasso is used to trace the trajectories of variables, 
 to detect the high value of the regularizing parameters, and detect changes in the regularization process,
 \item a double cross validation is used to select the best subset of variables for prediction based on the score,
 \item a GLM is used by the  best subset to predict via a simple cross validation.
\end{enumerate}
At last step the prediction accuracy of the method is calculated as average performance across hold-out predictions. 
For each observation $Y_i$,  the predicted value is $ \hat Y_i$. We assume that the observed value  $Y_i, 1 \leq i \leq n $ is really predict by
$ \hat Y_i$ if $-0.5 \leq Y_i - \hat Y_i \leq 0.5$. The prediction accuracy  $P_a$  is define by:
\begin{displaymath}     
      \left\{
      \begin{array}{l}
   P_a(\hat Y_i) = 1 \;\;\;\mbox{if}\; \;\; -0.5 \leq Y_i - \hat Y_i \leq 0.5 \; \cr
   P_a(\hat Y_i) = 0 \;\;\; \mbox{elsewhere}.
              \end{array}
                         \right.
      \end{displaymath}
      This calculation gives the number of  good predictions by each method and the power of prediction.
      The main quality criteria is the prediction power of a model and the other are
      the mean, the standard deviation, the quadratic risk  of prediction.
      For an any method we have :
      \begin{eqnarray*}
       \mbox{Mean}&=&\frac{1}{n}\sum_{i=1}^{n}\hat{Y}_i, \;\;\;  \mbox{Deviance} =Score(lambda.min), \;\;\; \mbox{Std}=\sqrt{\mbox{Deviance}}\cr       
       \mbox{Absolute risk}&=&\frac{1}{n}\sum_{i=1}^{n}|Y-\hat{Y}_i|,  \;\; \mbox{Prediction Power}  = \frac{100}{n}\times \# \{i, P_a(\hat Y_i) = 1, 1\leq i \leq n\}
      \end{eqnarray*}
where $\# A$ denote the cardinality of $A$
\subsection{Frequent variables}
Let $X=(X_1, X_2, \ldots, X_p)$ the set of all variables include interactions.
A each  step $j$ of  first level   of  LOLO-DCV, the Lasso provides 
the coefficients of all classes of variables and based on this  it can determine the presence or the absence
of each variable.
For any  $\lambda$, let  define the function "Presence" of variable like:
\begin{displaymath}     
     \left\{
      \begin{array}{l}
  \mathcal{P}_j(X_i)  = 1 \quad \mbox{if}\quad  \beta_i(\lambda) \neq 0\; \cr
  \mathcal{P}_j(X_i)  = 0 \quad  \mbox{elsewhere}.
              \end{array}
                         \right.
      \end{displaymath}
      where $\beta_i(\lambda)$ is a vector of coefficients of $X_i$ in the model at the step $j$.
      For a threshold $s, \quad  1\leq s \leq 100$, the set of frequent variables is 
      \begin{equation}
       \mbox{Var\_freq}(\lambda) = \left\{ X_i, \frac{100}{\max(j)}\times \sum_{l=1}^{max(j)}\mathcal{P}_j(X_i) \geq s\right\}
      \end{equation}
      \textbf{Notations}:\\
      $\mbox{Var freq lambda\_min} = \mbox{Var\_freq}(lambda.min)$\\
      $\mbox{Var freq lambda\_1se} = \mbox{Var\_freq}(lambda.1se)$\\
      LOLO DCV lambda$\_$min denote  LOLO-DCV using $\lambda  = lambda.min$\\ 
      LOLO DCV lambda$\_$1se denote  LOLO-DCV using $\lambda  = lambda.1se$
\subsection{interactions between variables}    
 In general experts in epidemiology and medicine decide to choose some interactions according to they knowledge and experience.
 To avoid this way of making, LOLO-DCV generate automatically all interactions in the full set of explanatory variables 
 used in model. This involves that the number of variables grows exponentially and the classical method of variable selection 
 will failed. LOLO-DCV  automatically  learn with all variables and all interactions and provides the optimal set of variables
for predictions.

\section{Results}
\subsection{Summary of results on prediction accuracy and quality criteria}
\begin{table}
\caption{
Summary of B-GLM prediction} \label{Summary_orig}
\begin{center}
\begin{tabular}{llllll}
\hline\noalign{\smallskip}
 Method & Mean & Deviance & Std & Absolute risk & Prediction Power (\%)\\
 \noalign{\smallskip}
\hline
\noalign{\smallskip}
B-GLM & 3.75 & 62.29 & 7.89 & 3.88 & 73.53\\
\hline
\end{tabular}
\end{center}
\end{table}
 \begin{table}
\caption{
Summary of original variables} 
\label{selection_variables_originales_sans_village_table}
\begin{center}
\begin{tabular}{llllll}
\hline\noalign{\smallskip}
 Method & Mean & Deviance & Std & Absolute risk & Prediction Power (\%)\\
 \noalign{\smallskip}
\hline
\noalign{\smallskip}
LOLO DCV lambda\_min & 3.75 & 72.04 & 8.49 & 4.48 & 78.76\\
LOLO DCV lambda\_1se & 3.75 & 72.04 & 8.49 & 4.48 & 78.76\\

Var freq lambda\_min & 3.75 & 68.05 & 8.25 & 3.96 & 74.84\\

Var freq lambda\_1se & 3.74 & 59.28 & 7.70 & 3.81 & 75.98\\
\hline
\end{tabular}
\end{center}
\end{table}

 \begin{table}
\caption{
Summary of original variables with village as fixed effect}  
\label{selection_variables_originales_avec_village_table}
\begin{center}
\begin{tabular}{llllll}
\hline\noalign{\smallskip}
 Method & Mean & Deviance & Std & Absolute risk & Prediction Power (\%)\\
\noalign{\smallskip}
\hline
\noalign{\smallskip}
LOLO DCV lambda\_min & 3.75 & 72.04 & 8.49 & 4.48 & 78.76\\

LOLO DCV lambda\_1se & 3.75 & 72.04 & 8.49 & 4.48 & 78.76\\

Var freq lambda\_min & 3.73 & 55.70 & 7.46 & 3.50 & 75.00\\

Var freq lambda\_1se & 3.74 & 57.33 & 7.57 & 3.63 & 76.80\\
\hline
\end{tabular}
\end{center}
\end{table}

 \begin{table}
\caption{
Summary of recoded variables} 
\label{selection_variables_recodee_sans_village_table}
\begin{center}
\begin{tabular}{llllll}
\hline\noalign{\smallskip}
 Method & Mean & Deviance & Std & Absolute risk & Prediction Power (\%)\\
\noalign{\smallskip}
\hline
\noalign{\smallskip}
LOLO DCV lambda\_min & 3.75 & 72.04 & 8.49 & 4.48 & 78.76\\

LOLO DCV lambda\_1se & 3.75 & 72.04 & 8.49 & 4.48 & 78.76\\

Var freq lambda\_min & 3.75 & 59.21 & 7.69 & 3.84 & 75.82\\

Var freq lambda\_1se & 3.74 & 59.97 & 7.74 & 3.81 & 73.86\\
\hline
\end{tabular}
\end{center}
\end{table}

 \begin{table}
\caption{
Summary of recoded variables with village as fixed effect} 
\label{selection_variables_recodee_avec_village_table}
\begin{center}
\begin{tabular}{llllll}
\hline
 Method & Mean & Deviance & Std & Absolute risk & Prediction Power (\%)\\
\noalign{\smallskip}
\hline
\noalign{\smallskip}
LOLO DCV lambda\_min & 3.75 & 72.04 & 8.49 & 4.48 & 78.76\\

LOLO DCV lambda\_1se & 3.75 & 72.04 & 8.49 & 4.48 & 78.76\\

Var freq lambda\_min & 3.73 & 60.11 & 7.75 & 3.87 & 76.31\\

Var freq lambda\_1se & 3.75 & 59.21 & 7.69 & 3.84 & 75.82\\
\hline
\end{tabular}
\end{center}
\end{table}
\subsection{Quality of estimators and optimal subset variables of prediction}

\textbf{Quality of estimators}
\begin{figure}
\begin{center}
\includegraphics[width=5in,height=3.4in]{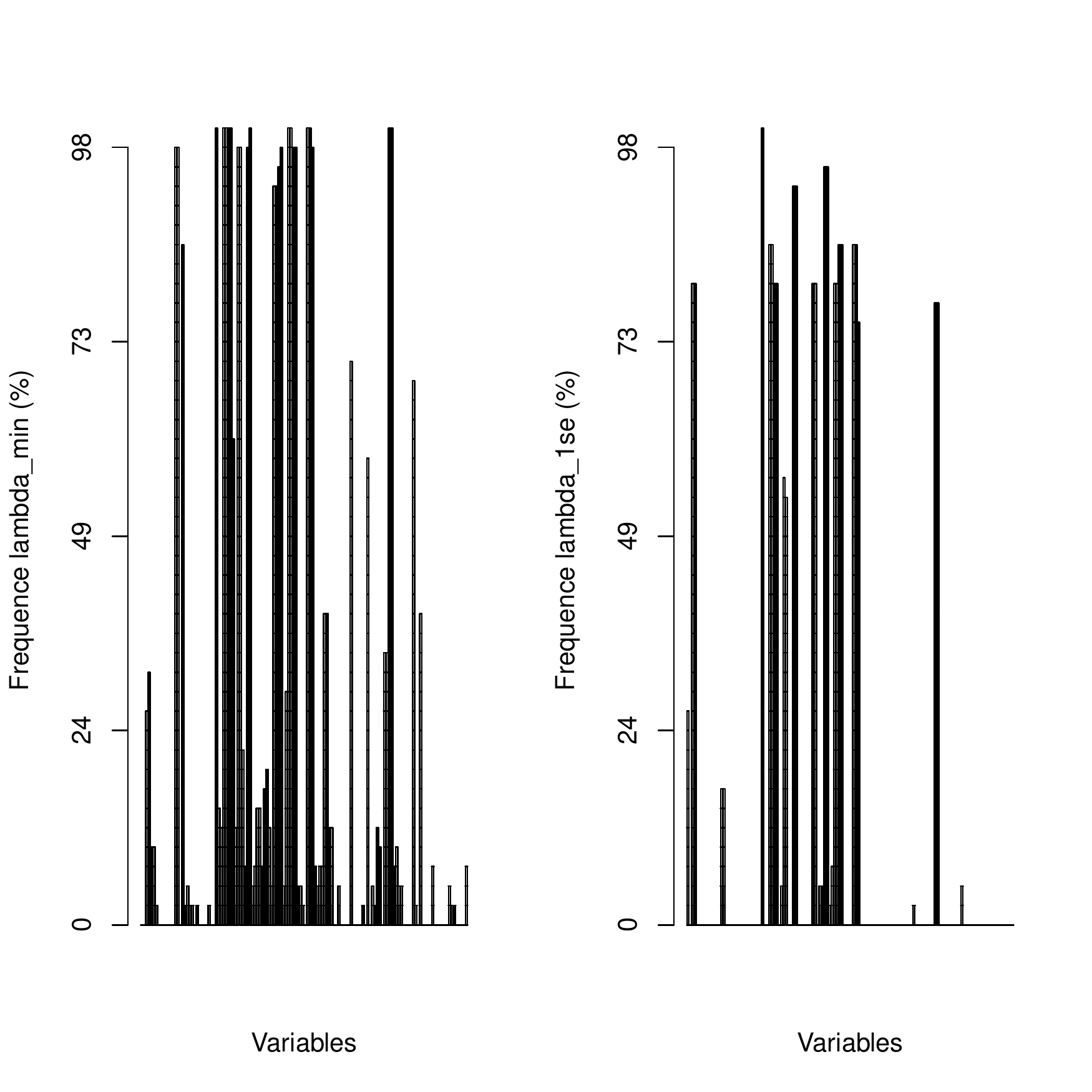}
\end{center}
\caption{
Frequent variables among original variables.
At the abscissas ($x_s$) are the variables include interactions and at the ordered $(y_s)$ the percentage of the presence of   variables}.
\label{selection_variables_originales_sans_village_graph}
\end{figure}
\begin{figure}
\begin{center}
\includegraphics[width=5in,height=3.3in]{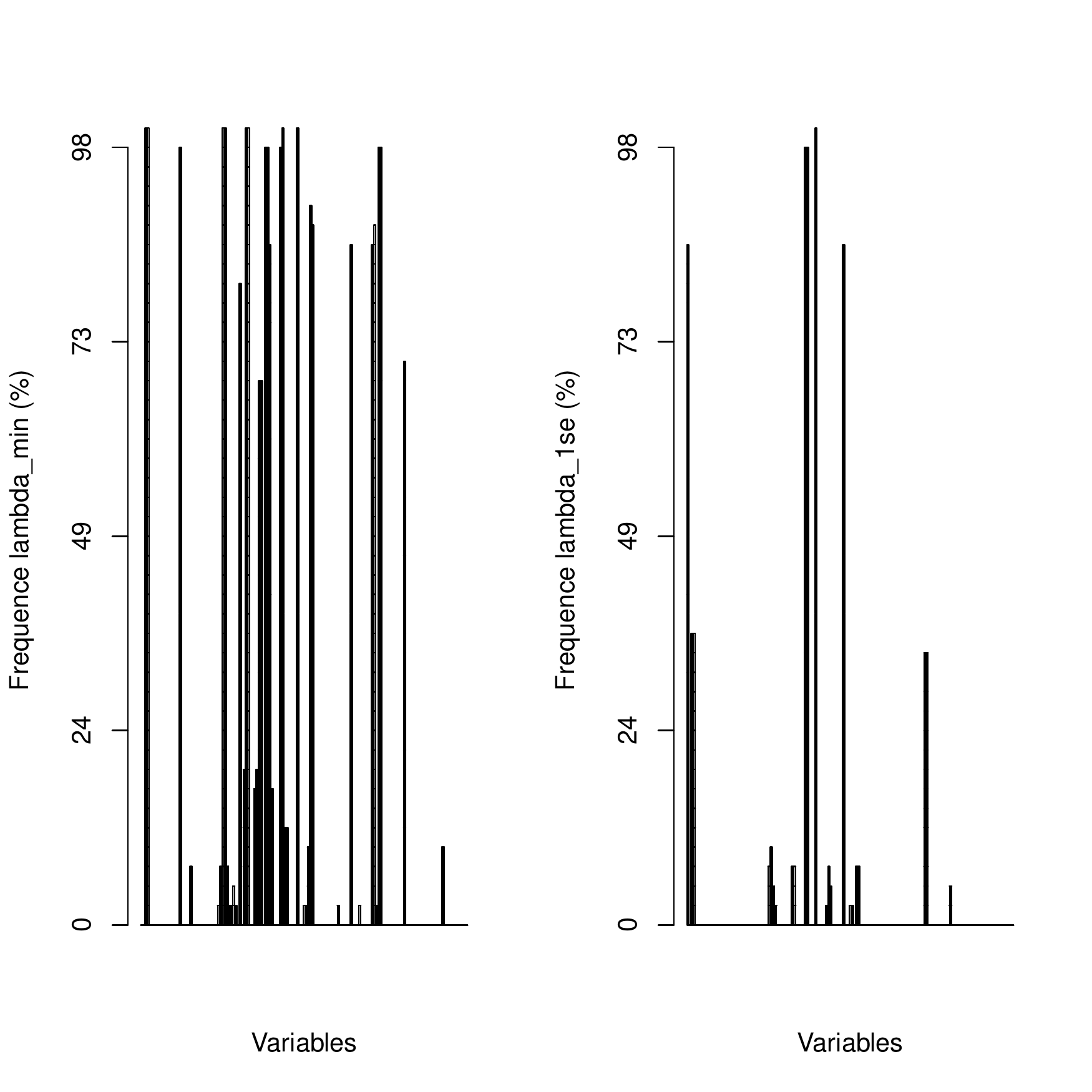}
\end{center}
\caption{
Frequent variables among original variables with village at fixed effect}
\label{selection_variables_originales_avec_village_graph}
\end{figure}
\textbf{Recoded variables}\\
\begin{figure}
\begin{center}
\includegraphics[width=5in,height=3.3in]{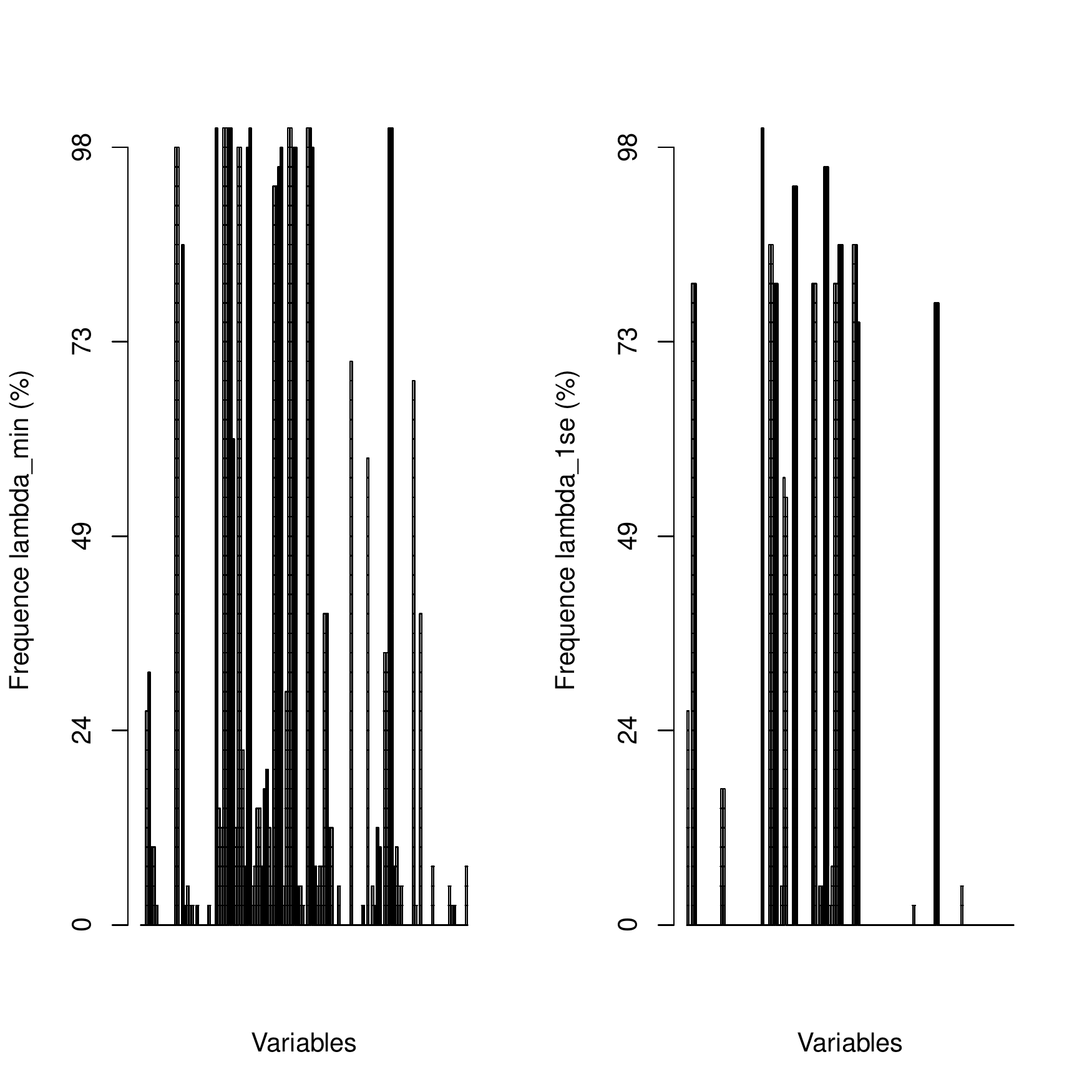}
\end{center}
\caption{
Frequent variables among recoded variables}
\label{selection_variables_recodee_sans_village_graph}
\end{figure}
\begin{figure}
\begin{center}
\includegraphics[width=5in,height=3.3in]{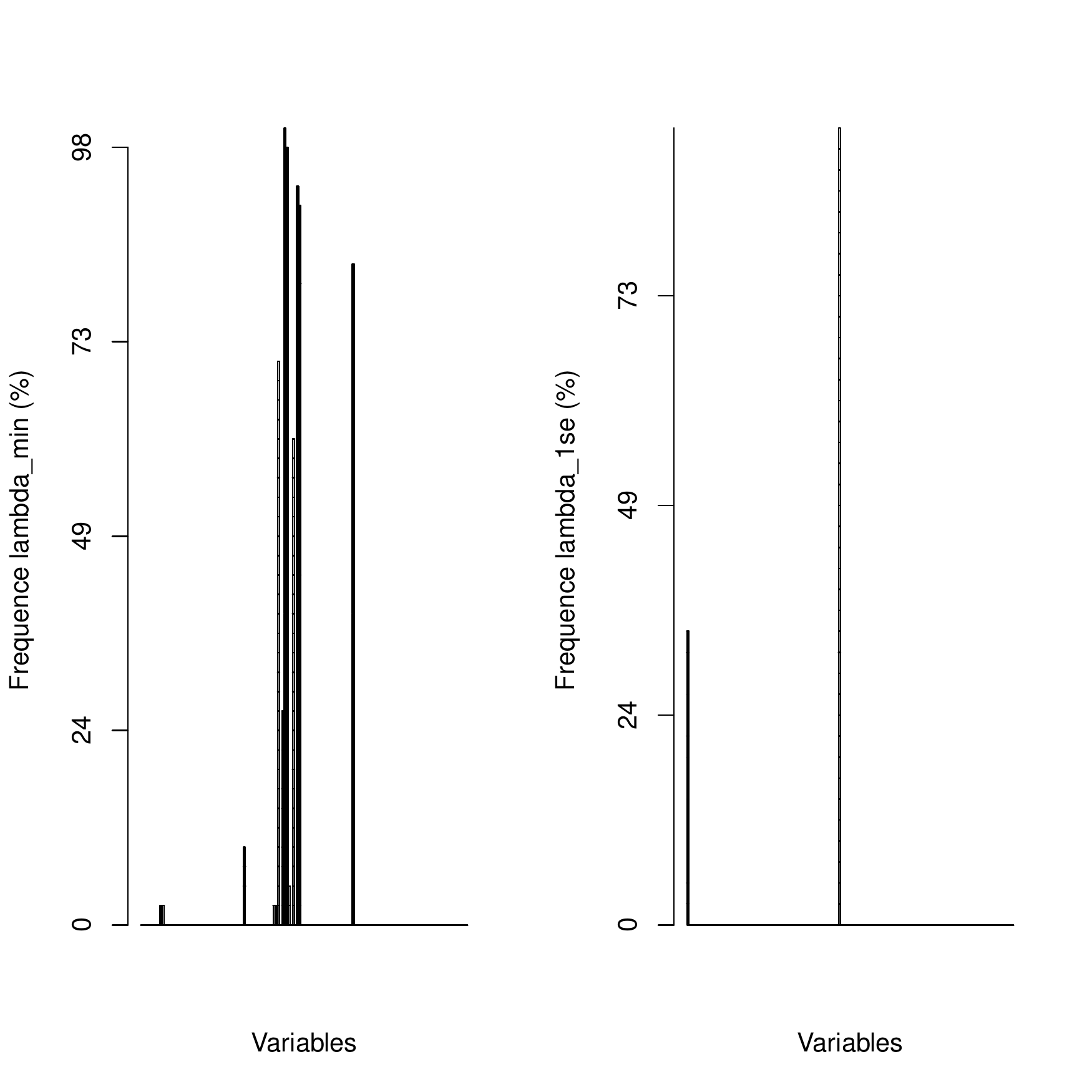}
\end{center}
\caption{
Frequent variables among recoded variables  with village at fixed effect }
\label{selection_variables_recodee_avec_village_graph}
\end{figure}

The best subset of variables selected by each method is:

\begin{enumerate}
 \item \textbf{B-GLM prediction}:\\
  Season (season), the number of rainy days during the three days of one survey (RainyDN),  
 mean rainfall between 2 survey (Rainfall), number of rainy days in the 10 days before the survey (RainyDN102), 
 the use of repellent (Repellent), The index of vegetation (NDVI)
the interaction between season and NDVI (season:NDVI) \cite{cottrell2012malaria}.
\item \textbf{LOLO-DCV}
\begin{enumerate}
 \item \textbf{Original variables}:\\
 Season and  interaction between  mean rainfall between 2 survey (season) and the number of rainy 
 days during the three days of one survey (Rainfall:RainyDN).

 \item \label{best subset}\textbf{Original variables with village as fixed effect}:\\
  Season (season) and interaction between number of rainy days in the 10 days before the survey and village (RainyDN10:village).
 \item \textbf{Recoded variables}:\\
  Season (season) and mean rainfall between 2 survey (RainyDN10).
 \item \textbf{ Recoded variables with village as fixed effect}:\\
 Season (season) and interaction between the number of rainy days during 
 the three days of one survey and presence of work around the site (RainyDN:Works).
\end{enumerate}
\end{enumerate}
LOLO-DCV$\_$lambda$\_$min and LOLO-DCV$\_$lambda$\_$1se achieves exactly the same performance in prediction: 
Mean, quadratic risk, absolute risk, and 
prediction power (Table. \ref{selection_variables_originales_sans_village_table}, 
\ref{selection_variables_originales_avec_village_table}, 
\ref{selection_variables_recodee_sans_village_table}, \ref{selection_variables_recodee_avec_village_table}).
The mean of predictions of the both methods is approximatively the same with the mean of observations (3.74) which  is achieved
exactly with frequent variables obtained
by lambda$\_$1se selection. \\
LOLO-DCV shows the Influence of interactions on the target variables.
The variability of the  score  in prediction
at village level (high in one the village (Dohinonko))
detect some problems in the data at this village. It  is confirmed by 
the experts. The   (Fig. \ref{selection_variables_originales_sans_village_graph}, \ref{selection_variables_originales_avec_village_graph}
, \ref{selection_variables_recodee_sans_village_graph}, \ref{selection_variables_recodee_avec_village_graph}) shows two class of variables, the most frequent and the least frequent.
The best Prediction power is with LOLO DCV (78.76) and the one with frequent variable is of LOLO-DCV lambda.1se (76.80), this
method also has the same mean in prediction with observations.
The number of variables and all interactions is $p=136$, the classical methods
will compute  $N=2^{136}$ different model before selecting the best subset.
Combined with   double cross-validation, calculation will be unrealizable  because 
of complexity of algorithm. The strength of LOLO-DCV is the usage of lasso and the two level cross validation. 
In a relative short time  LOLO-DCV detect all feature selected by the B-GLM
and some interpretable  interactions among them. The (Tables. \ref{Summary_orig}, \ref{selection_variables_originales_sans_village_table}, 
\ref{selection_variables_originales_avec_village_table}, 
\ref{selection_variables_recodee_sans_village_table}, \ref{selection_variables_recodee_avec_village_table}) show that
LOLO-DCV is relatively the best in prediction (mean, absolute risk, deviance) and the best with the prediction power. 
The distribution of the prediction error according to the classes of anopheles shows
a high  variability   for B-GLM 
 and low for the  LOLO-DCV. The optimal subset of feature 
obtained by LOLO-DCV algorithm is  approximatively the same at each step. This prove it's stability.
In final the best subset of variables for prediction is variables selected in
Original variables with village as fixed effect (\ref{best subset})

\section{Conclusion}
 Lasso  method combined with a GLM uses $L_1$ or  $L_2$ penalization on likelihood to estimate predictors. The
usage of  Lasso, GLM and  cross-validation for features selection is   an active area of research
in features and variables selection. In this work we propose LOLO-DCV method applied to a real data. 
The computation time is strongly  reduced with  the parallelization of the cross-validation loop.
The usage of levels for building the folds in cross-validation is important because a random sampling
 will use all characteristics of all levels and the predictions will be wrong. The experts can easily be helped
 by the machine in they decision. 
These results encourage a best  exploring of this approach for features selection. 
Adding random effects at some levels to improve our method is part of 
our future work possibly by combining the adaptative, group and sparse form of lasso procedure.

\section*{Annex}
\textbf{ Description of originales variables}
  \begin{table}[!ht]
\caption{
 Originale variables. \label{tab:Originales variables}}
\begin{center}
\begin{tabular}{llcl} 
\hline\noalign{\smallskip}
\textbf{Variables} &\textbf{Nature}&\textbf{Number of modalities}&\textbf{Modalities}\\
\noalign{\smallskip}
\hline
\noalign{\smallskip}
Repellent&Nominal& 2 & Yes/ No\\
Bed-net&Nominal &2 & Yes/  No\cr
Type of roof&Nominal& 2 & Tole/ Paille\cr
Ustensils& Nominal& 2 & Yes/  No\cr
Presence of constructions&Nominal& 2 & Yes/  No\cr
Type of soil &Nominal&2&Humid/ Dry\cr
Water course&Nominal&2&Yes/ No\cr
Majority Class&Nominal&3&1/4/7\cr
Season &Nominal&4&1/2/3/4\cr
Village&Nominal&9&\cr
House&Nominal&41&\cr
Rainy days before  mission &Numeric&Discrete&0/2/$\cdots$/9\cr
Rainy days during  mission &Numeric&Discrete&0/1/$\cdots$/3\cr
 Fragmentation Index &Numeric&Discrete&26/$\cdots$/71\cr
Openings &Numeric&Discrete&1/$\cdots$/5\cr
Number of inhabitants &Numeric&Discrete&1/$\cdots$/8\cr
Mean rainfall &Numeric&Continue&0/$\cdots$/82\cr
Vegetation&Numeric&Continue&115.2/$\cdots$/ 159.5\cr
Total Mosquitoes &Numeric&Discrete&0/$\cdots$/481\cr
Total Anopheles &Numeric&Discrete&0/$\cdots$/87\cr
Anopheles infected&Numeric&Discrete&0/$\cdots$/9\cr
\hline
\end{tabular}
\end{center}
\end{table}
\newpage
\textbf{Description of recoded variables}
  \begin{table}[!ht]
\caption{
Recoded variables. Variables with star are recoded. \label{tab:Recoded variables}}
\begin{center}
\begin{tabular}{llcl} 
\hline\noalign{\smallskip}
\textbf{Variables} &\textbf{Nature}&\textbf{Number of modalities}&\textbf{Modalities}\cr
\noalign{\smallskip}
\hline
\noalign{\smallskip}
Repellent&Nominal& 2 & Yes/ No\cr 
Bed-net&Nominal &2 & Yes/  No\cr
Type of roof&Nominal& 2 & Tole/ Paille\cr
Utensils& Nominal& 2 & Yes/  No\cr
Presence of constructions &Nominal& 2 & Yes/  No\cr
Type of soil &Nominal&2&Humid/ Dry\cr
Water course&Nominal&2&Yes/ No\cr
Majority class $^*$&Nominal&3&1/2/3\cr
Season &Nominal&4&1/2/3/4\cr
Village$^*$&Nominal&9&\cr
House $^*$&Nominal&41&\cr
Rainy days before  mission $^*$&Nominal&3&Quartile\cr
Rainy days during  mission &Numeric&Discrete&0/1/$\cdots$/3\cr
Fragmentation index $^*$&Nominal&4&Quartile\cr
Openings$^*$&Nominal&4&Quartile\cr
Nber of inhabitants $^*$&Nominal&3&Quartile\cr
Mean rainfall $^*$ &Nominal&4& Quartile\cr
Vegetation$^*$&Nominal&4&Quartile\cr
Total Mosquitoes &Numeric&Discrete&0/$\cdots$/481\cr
Total Anopheles&Numeric&Discrete&0/$\cdots$/87\cr
Anopheles infected &Numeric&Discrete&0/$\cdots$/9\cr
\hline
\end{tabular}
\end{center}
\end{table}
\newpage
\bibliography{Bibliography_MLDM_2015.bib}

\begin{thebibliography}{10}
\providecommand{\url}[1]{\texttt{#1}}
\providecommand{\urlprefix}{URL }

\bibitem{DBLP:conf/esann/Bontempi05}
Bontempi, G.: Structural feature selection for wrapper methods. In: {ESANN}
  2005, 13th European Symposium on Artificial Neural Networks, Bruges, Belgium,
  April 27-29, 2005, Proceedings. pp. 405--410 (2005),
  \url{https://www.elen.ucl.ac.be/Proceedings/esann/esannpdf/es2005-97.pdf}

\bibitem{cottrell2012malaria}
Cottrell, G., Kouway\`e, B., Pierrat, C., le~Port, A., Bouraïma, A., Fonton,
  N., Hounkonnou, M.N., Massougbodji, A., Corbel, V., Garcia, A.: Modeling the
  influence of local environmental factors on malaria transmission in benin and
  its implications for cohort study. PlosOne  7, ~8 (2012)

\bibitem{cross-validation1}
De~Bradanter, J., Pelckmans, K., Suykens, J.A.K., Vandewalle, J., De~Moor, B.:
  Robust cross validation score function with application to weigthed least
  squares support vector machine function estimation (2003), katholieke
  Universiteit Leuven, departement of electrical engineering, ESAT-SISTA

\bibitem{efron2004least}
Efron, B., Hastie, T., Johnstone, I., Tibshirani, R.: Least angle regression.
  The Annals of statistics  32(2),  407--499 (2005)

\bibitem{Friedman2010regularization}
Friedman, J., Hastie, T., Tibshirani, R.: Regularization paths for generalized
  linear models via coordinate descent. Journal of Statistical Software  33(1),
   1--22 (2010), \url{http://www.jstatsoft.org/v33/i01/}

\bibitem{goeman2010l1}
Goeman, J.J.: L1 penalized estimation in the cox proportional hazards model.
  Biometrical Journal  52(1),  70--84 (2010)

\bibitem{Guyon03anintroduction}
Guyon, I.: An introduction to variable and feature selection. Journal of
  Machine Learning Research  3,  1157--1182 (2003)

\bibitem{hastie2009element}
Hastie, T., Tibshirani, R., Friedman, J.: The Elements of Statistical Learning
  \textit{Data mining, Inference, Prediction}. Springer, second edn. (2009)

\bibitem{Friedman2015glmnet}
J.~Friedman, T.~Hastie, N.S., Tibshirani, R.: Lasso and elastic-net regularized
  generalized linear models (2015), \url{http://www.jstatsoft.org/v33/i01/} R
  CRAN

\bibitem{Ng97preventing"overfitting"}
Ng, A.Y.: Preventing "overfitting" of cross-validation data. In: Proceedings of
  the Fourteenth International Conference on Machine Learning. pp. 245--253.
  ICML '97, Morgan Kaufmann Publishers Inc., San Francisco, CA, USA (1997),
  \url{http://dl.acm.org/citation.cfm?id=645526.657119}

\bibitem{Osborne1}
Osborne, M., Presnell, B., Turlach, B.: A new approach to variable selection in
  least squares problems. IMA Journal of Numerocal Analysis  20,  389--403
  (2000)

\bibitem{tibshirani1996regression}
Tibshirani, R.: Regression shrinkage and selection via the lasso. Journal of
  the Royal Statistical Society. Series B (Methodological) pp. 267--288 (1996)

\bibitem{WHO}
WHO: World health organisation, world malaria report 2013, world global malaria
  programme. WHO Library Cataloguing-in-Publication Data p. 248 (2013)

\bibitem{zou2005regularization}
Zou, H., Hastie, T.: Regularization and variable selection via the elastic net.
  Journal of the Royal Statistical Society: Series B (Statistical Methodology)
  67(2),  301--320 (2005)

\end{thebibliography}
\end{document}